\documentclass[letterpaper]{article} 
\usepackage{aaai24}  
\usepackage{times}  
\usepackage{helvet}  
\usepackage{courier}  
\usepackage[hyphens]{url}  
\usepackage{graphicx} 
\usepackage{natbib}  
\usepackage{caption} 
\usepackage[namelimits]{amsmath} 
\usepackage{amssymb}             
\usepackage{amsfonts}            
\usepackage{mathrsfs}            
\usepackage{algorithm}
\usepackage{algorithmic}
\usepackage{newfloat}
\usepackage{listings}
\usepackage{color}
\usepackage{tabularx}
\usepackage{multirow}
\usepackage{multicol}
\usepackage{booktabs}
\usepackage{graphicx}
\usepackage{subcaption}
\usepackage[table,xcdraw]{xcolor}
\usepackage{bbding}
\usepackage{pifont}
\usepackage{wasysym}
\usepackage{comment}

\usepackage{algorithm}
\usepackage{algorithmic}
\usepackage{newfloat}
\usepackage{listings}

\usepackage{newfloat}
\usepackage{listings}
\DeclareCaptionStyle{ruled}{labelfont=normalfont,labelsep=colon,strut=off} 
\floatstyle{ruled}
\newfloat{listing}{tb}{lst}{}
\floatname{listing}{Listing}
\title{Dynamic Heterogeneous Federated Learning with Multi-Level Prototypes}
\author{
    Shunxin Guo\textsuperscript{\rm 1,2},
    Hongsong Wang\textsuperscript{\rm 1,2},
    Xin Geng$^{1,2}$\thanks{*Corresponding author}\\
}
\affiliations{
    \textsuperscript{\rm 1}MOE Key Laboratory of Computer Network and Information Integration, China\\
    \textsuperscript{\rm 2} School of Computer Science and Engineering, Southeast University, Nanjing 210096, China

}

\usepackage{bibentry}

\begin{document}
	
\maketitle

\begin{abstract}
Federated learning shows promise as a privacy-preserving collaborative learning technique. Existing heterogeneous federated learning mainly focuses on skewing the label distribution across clients. However, most approaches suffer from catastrophic forgetting and concept drift, mainly when the global distribution of all classes is extremely unbalanced and the data distribution of the client dynamically evolves over time. In this paper, we study the new task, i.e., Dynamic Heterogeneous Federated Learning (DHFL), which addresses the practical scenario where heterogeneous data distributions exist among different clients and dynamic tasks within the client. Accordingly, we propose a novel federated learning framework named Federated Multi-Level Prototypes (FedMLP) and design federated multi-level regularizations. To mitigate concept drift, we construct prototypes and semantic prototypes to provide fruitful generalization knowledge and ensure the continuity of prototype spaces. To maintain the model stability and consistency of convergence, three regularizations are introduced as training losses, i.e., prototype-based regularization, semantic prototype-based regularization, and federated inter-task regularization. Extensive experiments show that the proposed method achieves state-of-the-art performance in various settings.
\end{abstract}

\section{Introduction}
Federated learning (FL) has emerged as a significant research area that addresses the challenges of collaborative machine learning in decentralized and privacy-sensitive settings~\cite{mcmahan2017communication}.
Instead of centralizing data on a single server, FL models are trained locally on distributed devices such as smartphones to preserve data privacy.
Parameters from local models of clients are aggregated to create a global model that represents the collective knowledge of all participating edge devices.
This paradigm enables collaborative learning without compromising data privacy and security.
\begin{figure}[tb]
\begin{center}
   \includegraphics[width=1\linewidth]{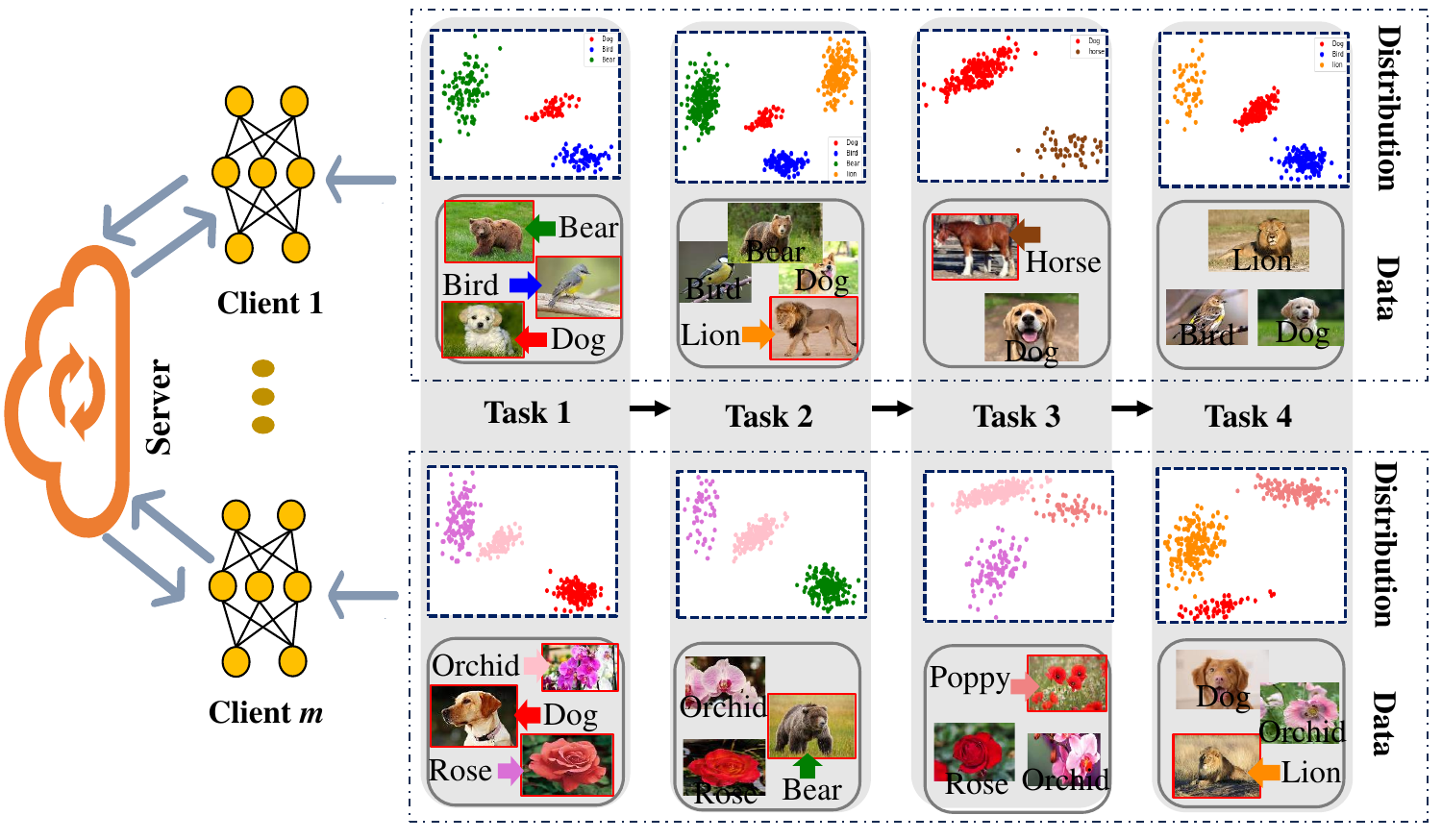}
\end{center}
   \caption{Dynamic Heterogeneous Federated Learning. Each participant incorporates dynamic data input from multiple tasks, where the class distribution varies between tasks. Images highlighted with red bounding boxes are objects of the newly introduced classes.}
\label{fig:intro}
\end{figure}

Data heterogeneity is an inevitable issue as different edge devices participating in collaborative learning collect data based on personal preferences.
There are significant differences in the underlying data distribution among the participants, resulting in their models being optimized according to local empirical risk minimization and further causing inconsistency in global optimization.
To address this problem, most previous methods~\cite{afonin2021towards,zhang2022fine,gao2022feddc,li2020federated} introduce global information to adjust the optimization direction of participants' local models, aiming to obtain a generalized global model.
Meanwhile, some other approaches~\cite{li2019fedmd,liang2020think,collins2021exploiting}
argue that it could be better for some clients to use their limited local data to train a personalized local model.
However, these approaches model in the static scenario of data heterogeneity and assume that the global class data is uniformly distributed while data within clients is fixed.

In practical scenarios, the data distribution is unbalanced or even long-tailed, and the client dynamically receives new data.
For instance, smartphones generate data throughout the day, and hospitals in the healthcare system periodically collect patient records. To accommodate this situation, we introduce a novel challenging problem, i.e., Dynamic Heterogeneous Federated Learning (DHFL), as illustrated in Figure~\ref{fig:intro}. For this task, there are diversified and unbalanced data distributions across clients and dynamic incremental tasks within the client.
The limited number of samples for rare classes leads to bias in both the local and aggregated global models.
The incremental arrival within the client causes \emph{local catastrophic forgetting}, i.e., local model updating to overfit the current data distribution and forget the knowledge of previous data distributions.
The aggregated model would quickly get into a local optimum and incur \emph{global concept drift} as the updated local model can be heavily biased toward majority classes in the current task distribution.
Existing FL methods typically perform poorly for this task due to a lack of consideration for prevalent long-tail and dynamic data distributions.

To address this problem, we introduce the Federated Multi-Level Prototype (FedMLP), which leverages intrinsic relations to build federated multi-level regularizations, enabling model stability training to achieve convergence consistency.
To mitigate the global concept drift exacerbated by the inconsistency of class distribution across clients and the imbalance of global class distribution, we introduce both prototype-based regularization and semantic prototype-based regularization.
These federated regularizations ensure that minority classes can leverage features from balanced semantic prototypes, thereby rectifying their own prototype representations. Consequently, local models are less biased towards dominant classes and show consistent performance in disadvantaged domains, thereby reducing the bias of the aggregated global classifier.
To address the local forgetting caused by dynamic incremental tasks within the client, we design the federated inter-task regularization, whose goal is to synchronize the recognized class prototypes of the current task with the class models of previous tasks. This regularization further reduces the bias introduced by inconsistent data samples and promotes feature similarity within the same class in different tasks.

The main contributions are summarized below.
\begin{itemize}
\item We study the novel and more practical problem, i.e., Dynamic Heterogeneous Federated Learning (DHFL), which assumes heterogeneous data distributions among clients and within them.
\item We construct the federated multi-level prototypes containing prototypes and semantic prototypes, which capture the correlation of underlying representations among samples and provide rich generalization information.
\item We develop federated multi-level regularizations during training that maintains model stability and consistency.
\end{itemize}

\section{Related Work}
\noindent{\textbf{Federated Learning with Data Heterogeneous.} Federated learning aims to facilitate collaborative model training across distributed devices or clients while preserving data locality.
The heterogeneity among edge devices introduces data heterogeneity, leading to the global concept drift and consequently affecting overall performance. Following the seminal work FedAvg~\cite{mcmahan2017communication}, which trains a global model by aggregating local model parameters, typical models such as FedProx~\cite{li2020federated}, pFedME~\cite{t2020personalized}, and FedDyn~\cite{durmus2021federated} mitigate the heterogeneity problem by relying primarily on the global penalty or aim to control the variance by calculating the global parameter stiffness.
Furthermore, since the statistical heterogeneity across clients or tasks is mainly concentrated in the labels, some other methods~\cite{collins2021exploiting, li2021ditto} propose learning shared data representations across clients. The MOON~\cite{li2021model}, FedProto~\cite{tan2022fedproto}, and FedProc~\cite{mu2023fedproc} aim to maximize feature-level consistency between local and global models.
These methods focus on resolving the data heterogeneity across clients while ignoring that, in reality, the global class distribution is extremely skewed or long-tailed, which would exacerbate the impact of this problem.
To address this issue, GRP-FED~\cite{2021GRP}, Ratio loss~\cite{afonin2021towards}, FEDIC~\cite{shang2022federated}, and CReFF~\cite{2022Federated} propose distillation using logits adjustment and calibration gate networks. However, these approaches require additional discriminators and public data, burdening participants or the server side. Moreover, they ignore that realistic federated learning environments typically involve multiple participants, each of which may experience different task increments and data changes.

\vspace{0.3mm}
\noindent{\textbf{Continual Federated Learning.} Catastrophic forgetting is a significant problem for continual federated learning. To alleviate this problem, existing works usually use the client to store the data of the old class internally~\cite{yang2021flop}, or apply the loss function to compensate for the forgetting of the old class, or extract the cross-task consistent inter-class relationship~\cite{dong2022federated}, or use an independent unlabeled proxy dataset~\cite{ma2022continual}.
However, these methods focus on the forgetting caused by the incremental learning of classes within the client and ignore the incremental learning of different tasks, i.e., the forgetting problem caused by the inconsistent class distribution between tasks.
The closely related method FedWeIT~\cite{yoon2021federated} uses client-side network parameter decomposition to reduce interference between different tasks. However, it involves cross-client migration, and each round of communication requires computational resources and time to handle weight adjustment and information transfer, resulting in increased training time.
The FedNTD~\cite{lee2022preservation} uses the global model to predict the locally available data to solve the forgetting caused by the data heterogeneity but requires the global model to be unbiased to avoid similar trends in the trained local model. In this paper, we introduce a consensus optimization of global semantic prototypes, i.e., knowledge complements for similarity classes and local prior prototype knowledge to learn a generalizable global model as well as robust and stable local models during federated learning.

\vspace{0.3mm}
\noindent{\textbf{Cluster-Based Federated Learning.} Existing cluster-based federated learning~\cite{taik2022clustered} involves grouping clients with similar data distributions into clusters and uniquely associating each client with a specific data distribution based on model parameters~\cite{long2023multi}, gradient information~\cite{sattler2020clustered}, and training loss~\cite{li2021model}. These approaches facilitate training customized models tailored to each distribution~\cite{caldarola2021cluster}. The CPCL~\cite{huang2023rethinking} uses clustering strategies to select representative prototypes and addresses the FL problem with the domain transfer. In our approach, we use prototype clustering to mine relationships between minority classes and their similar classes, thereby addressing the bias in the global classifier. In addition, we perform initialization for newly introduced classes to reduce the introduced bias and enable faster convergence to their optimal representation.

\section{Dynamic Heterogeneous Federated Learning}
\begin{figure*}[!ht]
	\begin{center}
		\includegraphics[width=1\linewidth]{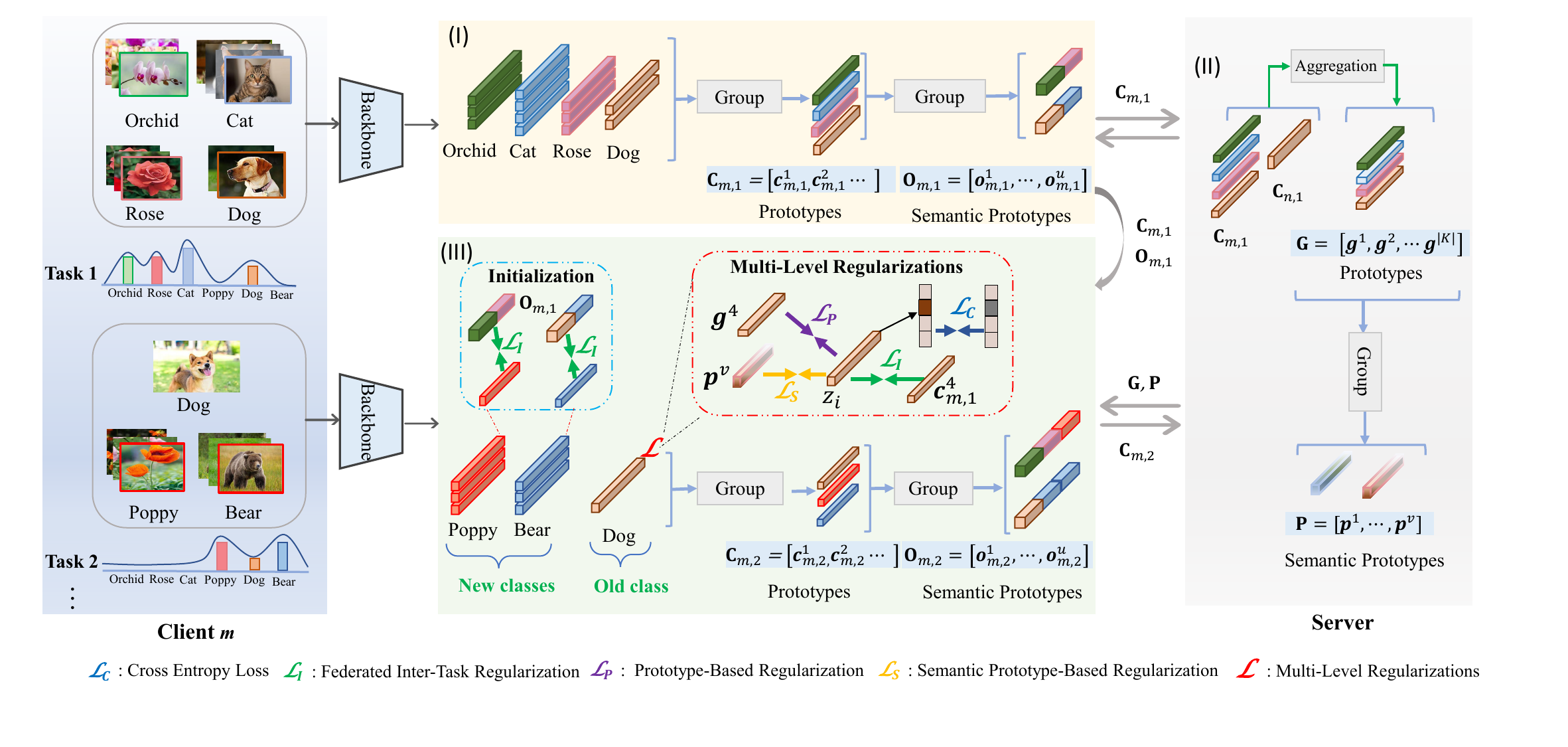}
	\end{center}
	\caption{Illustration of FedMLP. Participants generate prototypes and upload them to the server. Both the participants and the server group the prototypes to obtain semantic prototypes. For new task training within participants, we construct \textbf{Multi-Level Regularizations} including \textit{prototype-based regularization} to maintain the global prototype space consistency, \textit{semantic prototype-based regularization} to supply generalization information, and \textit{federated inter-task regularization} to provide stability and locally consistent continuity signals. New classes are highlighted in red.}
	\label{fig:method}
 \vspace{-3mm}
\end{figure*}

\subsection{Problem Formulation}
For dynamic heterogeneous federated learning (DHFL), there exists a central server and $M$ participants, and each participant is associated with $T$ stages of tasks. The private data of the participant is incrementally added in an online manner, defined as $D_{m} = \{D_{m,t}\}_t^T$, where $m$ is the client index. The $t$-th task corresponding private data denoted as $D_{m,t}=\left\{(x_{i},y_{i})\right\}_{i=1}^{N_{m,t}}$ consists of $N_{m,t}$ pairs of samples $x_{i}$, whose corresponding label is $y_i \in \mathbb{R}^{{|K|}\times 1}$, where $K$ is defined as the set of classes for classification. Due to the varied and skewed distribution of data across tasks and participants, \emph{local catastrophic forgetting} and \emph{global concept drift} are two critical problems for DHFL.

\noindent{\textbf{Local Catastrophic Forgetting.}} Since $P_{m,t}(y) \neq P_{m,t+1}(y)$, there exists local model forgetting caused by class distribution inconsistent within the local client. If the new task contains novel classes that have not been trained, the local federated model focuses on the parameters associated with them and forgets the parameters of the old classes.

\noindent{\textbf{Global Concept Drift.}} Since $P_{m,t}(y) \neq P_{n,t}(y)$ and $P_{m,t}(y | x) \neq P_{n,t}(y | x)$, there exists a skewed class distribution in that the marginal distributions $P(y)$ vary across clients of the same stage, which makes the local classifiers fit their respective distributions and leads to a global classifier shift after aggregation.  Furthermore, the skewed class distribution, i.e., $P(y_i) \neq P(y_j)$, exacerbates this issue as the classifier tends to be biased toward classes with larger number samples.

\subsection{Method Overview}
For simplicity, we assume that the model architectures of different participants are identical and that the model consists of two modular components: the feature extractor and the unified classifier.
The feature extractor denoted as $f: \mathcal{X} \rightarrow \mathcal{Z}$, encodes the sample $x_i$ into a $d$ dimension feature vector $z_i = f(x_i) \in \mathbb{R}^{d} $ within the feature space $\mathcal{Z}$.
A unified classifier denoted as $l_i = g(z_i) \in \mathbb{R}^{|K|}$, maps features $z_i$ to logits output $l_i$. Applying the softmax function to $l_i$ yields the predicted probability distribution $\hat{y}_i$.

The framework of the proposed method is illustrated in Figure~\ref{fig:method}.
To obtain stable consistency information and fruitful generalization knowledge, \textit{Federated Multi-Level Prototypes} are constructed in the participants and the server.
To achieve unbiased models and continuous feature space, \textit{Federated Multi-Level Regularizations} are introduced during local updating.

\subsection{Federated Multi-Level Prototypes}
\subsubsection{Federated Prototype.}
Each prototype represents the average of feature vectors corresponding to the same class and encapsulates distinctive semantic information.
In the DHFL scenario, there is an inconsistency of prototypes across participants due to their distinctive local models and private data. This suggests that prototypes inherently convey participant-specific information, which inspires us to use prototypes from different sources to train the generalizable model without compromising privacy. We define the prototype of the $k$-th class from the $m$-th client at the $t$-th stage as follows:
\begin{equation}
\begin{split}
\label{fig:proto}
\mathbf{\mathit{c}}_{m, t}^{k}=\frac{1}{\left|D_{m, t}^{k}\right|} \sum_{x_{i}\in D_{m, t}^{k}} f_{m}\left(x_{i}\right),
\end{split}
\end{equation}
where $D_{m,t}^{k}=\left\{x_{i},y_i \mid y_{i}=k\right\} \subset D_{m,t}$ denotes the set of samples belong to the $k$-th class within the $m$-th client on the $t$-th stage.
The local prototype set of the $t$-th stage task on the $m$-th participant degree can be defined as:
\begin{equation}
\begin{split}
\mathbf{C}_{m,t}=\left[\mathbf{\mathit{c}}_{m,t}^{1}, \ldots, \mathbf{\mathit{c}}_{m,t}^{k}\right],
\end{split}
\end{equation}
where $\mathbf{C}_{m,t}  \in \mathbb{R}^{{|K_{m,t}|} \times d}$ and $K_{m,t}$ indicates the set of classification classes for the $t$-th stage task on $m$-th participant.

Given that the federated learning training phase involves numerous devices, communication efficiency becomes crucial. A straightforward solution using prototype learning is to aggregate global class prototypes on the server side via averaging.
The aggregation of the $k$-th class prototype is defined as:
\begin{equation}
\label{split}
\begin{split}
\mathbf{\mathit{g}}^{k}=\frac{1}{M^k} \sum_{m=1}^{M^k} \mathbf{\mathit{c}}_{m,t}^k,
\end{split}
\end{equation}
where $M^k$ represents the number of participants contains $k$-th class.
After multiple rounds of collaborative learning, the global prototype of all classes can be obtained as:
\begin{equation}
\begin{split}
\mathbf{G}=\left[\mathbf{\mathit{g}}^{1}, \ldots, \mathbf{\mathit{g}}^{k},\ldots, \mathbf{\mathit{g}}^{|K|}\right],
\end{split}
\end{equation}
where $\mathbf{G} \in \mathbb{R}^{{|K|} \times d}$ contains prototypes of all classes.

\subsubsection{Federated Semantic Prototype.}
We utilize an unsupervised clustering algorithm to select a set of representative semantic prototypes as supplementary knowledge for prototypes, with each semantic prototype conforming to a similar distribution of object semantics. The federated semantic prototype could eliminate the influence of heterogeneous data and enhance generalization and discriminative capabilities.

The federated semantic prototype effectively initializes new class prototypes of new tasks according to the similar continuity of semantic space. It reduces the difference in class distribution between tasks to solve the problem of local model forgetting. The local semantic prototype used for initialization is obtained by clustering the existing prototypes $\mathbf{C}_{m,t}$, which is defined as $\mathbf{O}_{m,t} = \left[\mathbf{\mathit{o}}^1, \ldots, \mathbf{\mathit{o}}^u\right]$, where $\mathbf{\mathit{o}}^u$ represents the $u$-th local semantic prototype.

Considering global classifier shifts due to class distribution differences across participants and unbalanced global class distributions, we also cluster global prototypes $\mathbf{G}$ to capture inter-class similarities without being specific to individual participants. The obtained global semantic prototype can be used as the supplementary information of the minority class, which is defined as $\mathbf{P} =  \left[ \mathbf{\mathit{p}}^1, \ldots, \mathbf{\mathit{p}}^v\right]$, where $\mathbf{\mathit{p}}^v$ represents the $v$-th global semantic prototype.

\subsection{Federated Multi-Level Regularizations}
To alleviate the problems of \emph{local catastrophic forgetting} and \emph{global concept drift}, we propose federated multi-level regularizations as the training objectives, mainly consisting of the following three terms.
\subsubsection{Prototype-Based Regularization.}
To learn the shared information of classes, the prototype-based regularization is designed to encourage the feature vector $z^k_i$ generated by the local model for the query sample $x^k_i$ of the $k$-th class to approximate their corresponding global class prototypes.
The commonly mean squared error loss may excessively focus on features with larger differences and less generality, potentially leading to gradient explosion~\cite{zhang2022residual}. To avoid this issue, we adopt the Smooth L1 loss~\cite{girshick2015fast}, and this regularization $\mathcal{L}_{P}$ is as follows:
\begin{equation}
\label{GP}
\mathcal{L}_{P} = \frac{1}{N_{m,t}}\sum_{k \in K_{m,t}}\sum_{i=1}^{|D^k_{m,t}|}\mathrm{smooth}_{\mathit{L_1}}(z_i^k, \mathbf{\mathit{g}}^{k}),\\
\end{equation}
where $\mathbf{\mathit{g}}^{k} \in \mathbf{G}$ denotes the $k$-th trained global class prototype. As Eq.~\eqref{GP}, the purpose is to learn prototype-related knowledge from others while preserving privacy, so as to ensure that similar classes are closer to each other, while dissimilar classes are relatively far apart. This helps to better organize and distinguish different classes, thus improving model classification performance.

\subsubsection{Semantic Prototype-Based Regularization.}
For the minority classes, we further optimize using semantic prototypes of the corresponding clusters:
\begin{equation}
\label{gpsf}
\mathcal{L}_{S} = \frac{1}{|D^k_{m,t}|}\sum_{i=1}^{|D^k_{m,t}|}\sum_{k \in J}\mathrm{smooth}_{\mathit{L_1}}(z_i^k,\mathbf{\mathit{p}}^{v}),
\end{equation}
where $\mathbf{\mathit{p}}^{v} \in \mathbf{P}$ is the semantic prototype of the cluster corresponding to the $k$-th class, and $J$ denotes the set of minority classes.
This loss enables the minority classes with limited sample sizes to learn similar feature representations alleviating the global classifier shift.


\subsubsection{Federated Inter-Task Regularization.}
To address the problem of local catastrophic forgetting, the primary goal is to ensure that prototypes of old classes do not exhibit significant biases in the current task while effectively training prototypes of new classes.
For newly introduced classes, we deem that the initialization of a well-generalized representation should not only have clear decision boundaries but also be as general as possible to the similar old class feature space. Thus, it can provide a stable direction of convergence to quickly adapt the model to new tasks.

The federated inter-task regularization $\mathcal{L}_{I}$ is devised to promote consistency and continuity of the feature space in the current and previous tasks to alleviate local forgetting and achieve stable convergence.
Similar to~\cite{yang2022enhancing}, we use the Kullback-Leibler divergence and define this regularization as:
\begin{equation}
\label{inter}
\mathcal{L}_{I} = \frac{1}{N_{m,t}}\sum_{k \in K_{m,t}}\sum_{i=1}^{|D^k_{m,t}|} \mathit{KL} \left ( z_{i}^{k} \parallel \tilde{\mathit{c}}_{m,t-1}^{k}\right ),
\end{equation}
where $\tilde{\mathit{c}}_{m,t-1}^{k}$ is defined as:
\begin{equation}
\tilde{\mathit{c}}_{m,t-1}^{k} = \left\{\begin{array}{ll}
\mathbf{\mathit{c}}_{m,t-1}^{k}, &\text{if}\quad \mathbf{\mathit{c}}_{m,t-1}^{k} \in \mathbf{C}_{m,t-1}, \\
\mathbf{\mathit{c}}^n, &\text{otherwise },
\end{array}\right.
\end{equation}
where $\mathbf{\mathit{c}}_{m,t-1}^{k}$ represents the old prototype corresponding to the class $k$, and $\mathbf{\mathit{c}}^n$ is the initialization representation for the new class, defined as follows:
\begin{equation}
\mathbf{\mathit{c}}^n =
\underset{\mathbf{\mathit{o}}^u \in \mathbf{O}_{m, t}}{\arg \min }\left\|z_{i}^{k}-\mathbf{\mathit{o}}^u\right\|_{2}.
\end{equation}

Note that minimizing Eq.~\eqref{inter} could promote consistency between the embedding vectors and the local prototypes of the old classes. This ensures that the new classes do not deviate significantly from the overall class distribution, preventing them from dominating the training process or causing the model to converge in a biased direction.

\subsection{Training Objective}
To maintain the local discriminative ability, the supervision Cross Entropy loss~\cite{de2005tutorial} is used as classification loss, which is shown below:
\begin{equation}
\label{ce}
\mathcal{L}_{C} = -\frac{1}{N_{m,t}}\sum_{i=1}^{N_{m,t}}y_i \cdot \log(\hat{y}_i),
\end{equation}
where $\hat{y}_i$ denotes the predicted probability distribution of the $i$-th sample.
Thus, the objective of optimization during the local training update is:
\begin{equation}
\label{all}
\mathcal{L}= \mathcal{L}_{C} + \mathcal{L}_{P} +\mathcal{L}_{I} + \alpha\mathcal{L}_{S},
\end{equation}
where $\alpha$ is a hyperparameter to decide whether to learn other similar global prototypes.
During the local update, the local data is optimized guided by the objective defined in Eq.~\eqref{all} based on the multi-level prototype knowledge.

FedMLP uploads only shallow parameters focusing on local feature extraction to the collaborative learning server. This reduces communication overhead and improves privacy. Compared with existing methods that utilize the global model to construct regularization terms, the size of the prototype is much smaller than the model parameters, which reduces the computational cost for participants. In addition, model updates use prototype knowledge stored from previous tasks without compromising privacy. The semantic prototype remains privacy-preserving by undergoing two averaging operations~\cite{tan2022fedproto}.

\begin{table*}[tb]
\centering
\scalebox{1}{
\begin{tabular*}{17cm}{@{\extracolsep{\fill}}ccc|cc|cc|cc|cc }
\toprule
\cmidrule{1-11}
& \multicolumn{2}{c}{} & \multicolumn{2}{c}{}   & \multicolumn{4}{c}{\textbf{CIFAR-10}}& \multicolumn{2}{c}{} \\
& \multicolumn{2}{c}{\multirow{-2}{*}{\textbf{MNIST}}} & \multicolumn{2}{c}{\multirow{-2}{*}{\textbf{FEMNIST}}} & \multicolumn{2}{c}{\textit{s} = 2}& \multicolumn{2}{c}{\textit{s} = 4} & \multicolumn{2}{c}{\multirow{-2}{*}{\textbf{CIFAR-100}}}\\
\cmidrule{2-11}
\multirow{-3}{*}{\textbf{Method}}& $\mathcal{A}^{loc}$  & $\mathcal{A}^{glo}$ & $\mathcal{A}^{loc}$ & $\mathcal{A}^{glo}$  & $\mathcal{A}^{loc}$ & $\mathcal{A}^{glo}$  & $\mathcal{A}^{loc}$ & $\mathcal{A}^{glo}$ & $\mathcal{A}^{loc}$ & $\mathcal{A}^{glo}$\\
\midrule
    FedAvg  & 98.36 & 98.35 & 85.95  & 77.32  & 70.57 & 72.51 & 73.52 & 74.46 & 54.01 & 37.32\\
    FedProx & 98.66 & 98.55 & \underline{89.58}  & 80.19  & 79.03 & 74.68 & 78.50 & 76.21 & 60.29 & 38.12\\
    FedRep  & \underline{98.69} & \underline{98.71} & 89.15  & 81.11  & \underline{79.73} & \underline{77.25} & \underline{79.74} & \underline{77.66} & \underline{60.52} & 42.96\\
    FedProto  & 98.24 & 98.17 & 86.05  & \underline{85.56}  & 68.10 & 68.25 & 68.14 & 68.25 & 57.84 & \underline{57.09}\\
    FedNTD  & 98.37 & 98.16 & 75.96  & 40.79  & 78.34 & 74.26 & 78.47 & 76.65 & 53.73 & 37.03\\
    CreFF   & 90.07 & 97.92 & 34.80  & 64.61  & 29.77 & 55.51 & 42.65 & 69.48 & 16.11 & 29.40\\
    \midrule
    FedMLP   & \textbf{98.76} & \textbf{98.91} & \textbf{89.91}  & \textbf{89.03}  & \textbf{80.84} & \textbf{82.98} & \textbf{80.56} & \textbf{81.96} & \textbf{62.33} & \textbf{61.02}\\
\bottomrule
\end{tabular*}}
  \caption{Comparison with state-of-the-art methods under \textit{Sharding} non-iid partition strategy with $\gamma$ = 0.5 and $T$ = 5. Note that we highlight the \textbf{best} results in bold and the \underline{second best} results in underlining.}
\label{tab: com}
\end{table*}

\begin{table*}[tb]
\centering 
\scalebox{1}{
\begin{tabular*}{17cm}{@{\extracolsep{\fill}}ccc|cc|cc|cc|cc }
\toprule
\cmidrule{1-11}
& \multicolumn{2}{c}{} & \multicolumn{2}{c}{}   & \multicolumn{4}{c}{\textbf{CIFAR-10}}& \multicolumn{2}{c}{} \\
& \multicolumn{2}{c}{\multirow{-2}{*}{\textbf{MNIST}}} & \multicolumn{2}{c}{\multirow{-2}{*}{\textbf{FEMNIST}}} & \multicolumn{2}{c}{$\beta$ = 0.1}& \multicolumn{2}{c}{$\beta$ = 1.0} & \multicolumn{2}{c}{\multirow{-2}{*}{\textbf{CIFAR-100}}}\\
\cmidrule{2-11}
\multirow{-3}{*}{\textbf{Method}}& $\mathcal{A}^{loc}$  & $\mathcal{A}^{glo}$ & $\mathcal{A}^{loc}$ & $\mathcal{A}^{glo}$  & $\mathcal{A}^{loc}$ & $\mathcal{A}^{glo}$  & $\mathcal{A}^{loc}$ & $\mathcal{A}^{glo}$ & $\mathcal{A}^{loc}$ & $\mathcal{A}^{glo}$\\
\midrule
FedAvg   & 97.54 & 97.81 & \underline{72.62} & \underline{76.81} & 71.38 & 60.05 & \underline{75.64} & 73.92 & 39.44 & \underline{42.23} \\
FedProx  & \underline{98.08} & \underline{98.44} & 71.26 & 74.44 & 70.99 & 56.66 & 74.03 & 71.64 & 38.69 & 40.53 \\
FedRep   & 97.91 & 98.31 & 69.41 & 73.48 & 63.42 & 64.89 & 72.42 & \underline{74.10} & \underline{40.18} & 42.01 \\
FedProto & 96.07 & 96.54 & 38.94 & 33.50 & 66.86 & \underline{68.61} & 58.29 & 58.77 & 17.33 & 17.06 \\
FedNTD   & 97.59 & 97.66 & 46.01 & 50.26 & \underline{72.25} & 51.65 & 75.55 & 70.14 & 38.49 & 39.15 \\
CreFF    & 91.07 & 97.15 & 35.11 & 42.12 & 41.55 & 52.55 & 39.92 & 62.85 & 15.28 & 26.02 \\
\midrule
FedMLP    & \textbf{98.16} & \textbf{98.56} & \textbf{74.66 }& \textbf{77.36} & \textbf{79.26} & \textbf{78.41} & \textbf{77.86} & \textbf{78.89} & \textbf{42.27} &\textbf{ 42.57} \\
\bottomrule
\end{tabular*}}
\caption{Comparison with state-of-the-art methods under \textit{DDA} non-iid partition strategy.}
\label{tab:com2}
\end{table*}

\vspace{-1mm}
\section{Experiments}
\subsubsection{Datasets.}
We evaluate the methods on four benchmark datasets: CIFAR-10~\cite{krizhevsky2009learning}, CIFAR-100~\cite{krizhevsky2009learning}, MNIST~\cite{deng2012mnist}, FEMNIST~\cite{cohen2017emnist,caldas2018leaf}. Notably, the FEMNIST dataset contains 62 classes, each containing 600 samples.
For fair comparisons with existing works, we use a 5-layer CNN model~\cite{krizhevsky2012imagenet} for the MNIST datasets and utilize the ResNet-18 model~\cite{He2016Deep} for the CIFAR datasets.


We follow the approach of~\cite{li2020federated} and use Top-1 accuracy for fair evaluation during the testing phase. We take the accuracy average over the last ten communication epochs as the final performance. To verify whether the proposed method mitigates the forgetting problem when maintaining continual learning, the test set of the current task includes the corresponding test sets of all previous tasks. In addition, a balanced test set containing all samples is used to evaluate both the private models of individual clients and the global model. We denote the accuracy on the corresponding class set for each client's private model as $\mathcal{A}^{sel}$, the accuracy on the balanced test set as $\mathcal{A}^{loc}$, and the accuracy on the balanced test set for the global model as $\mathcal{A}^{glo}$.
\subsubsection{Implementation Details.}
We divide the data into 100 tasks with 20 clients and randomly select 10 participants in each round, i.e., $M$ = 10. To ensure fairness across clients, all clients contain the same number of fixed tasks, i.e., $T$ = 5. Additionally, to simulate privacy within clients, the tasks included in each client are predetermined.

We conduct 50 global communication epochs (each epoch globally contains $T$ interactions) and 20 rounds of local updates, where all FL methods show little to no accuracy gain from increased communication. We use an SGD optimizer with a learning rate of 0.01, a weight decay of 1$e$-5, and a momentum of 0.9. The training batch size is set to 32.
\vspace{-1mm}
\subsection{Experimental Setup}
\subsubsection{Experimental Settings.}
As the tasks are data heterogeneity, i.e., non-independent and non-identically distributed (non-iid), we employ two distinct partitioning strategies:
\begin{itemize}
      \item \textit{Sharding}~\cite{collins2021exploiting}. This setting sorts all training sample data by label, divides the data into shards of equal size according to the number of tasks, and controls heterogeneity by $s$, i.e., the number of shards in each task. The data size of each stage task is the same, but the class distribution is different in \textit{Sharding}. We set $s$ to MNIST ($s$ = 4), FEMNIST ($s$ = 4), CIFAR-10 ($s \in \{2, 4\}$) and CIFAR-100 ($s$ = 4).
    \item \textit{Dirichlet Distribution Allocation (DDA)}~\cite{luo2021no,lee2022preservation}. This setting assigns a partition of class $k$ by sampling $P_k \approx Dir(\beta)$ according to the Dirichlet distribution. The distribution and dataset size are different for each task in \textit{DDA}. We set $\beta$ to MNIST ($\beta$ = 1.0), FEMNIST ($\beta$ = 1.0), CIFAR-10 ($\beta \in \{1.0, 0.1\}$) and CIFAR-100 ($\beta$ = 1.0).
\end{itemize}

To construct the long-tailed distribution of the global class data, we follow ~\cite{cao2019learning} to use the imbalance rate to represent the ratio between the sample sizes of the least frequent class and the most frequent class, i.e., $\gamma =\min_{i}\left\{n_{i}\right\}/ \max_{i}\left\{n_{i}\right\}$. We set $\gamma \in \{1, 0.5, 0.05\}$. Specifically, we define minority classes as those residing in the lower 50\% of sample sizes within the dataset, indicating that these classes have relatively fewer samples. The partitioning of the test sets on each client is based on the labels corresponding to the training sets, enabling different samples for the same classes across clients.

\subsubsection{Comparison Baselines.}
We compare our method with representative state-of-the-art FL methods: FedAvg~\cite{mcmahan2017communication,collins2021exploiting}, FedProx~\cite{li2020federated}, FedRep~\cite{collins2021exploiting}, FedProto~\cite{tan2022fedproto}, FedNTD~\cite{lee2022preservation} and CReFF~\cite{2022Federated}.
FedNTD~\cite{lee2022preservation} mitigates heterogeneity in data by addressing catastrophic forgetting, and CReFF~\cite{2022Federated} addresses the long-tail distribution of global classes.

\begin{figure}[tb]
    \centering
    \begin{subfigure}{0.22\textwidth}
        \centering
        \includegraphics[width=\linewidth]{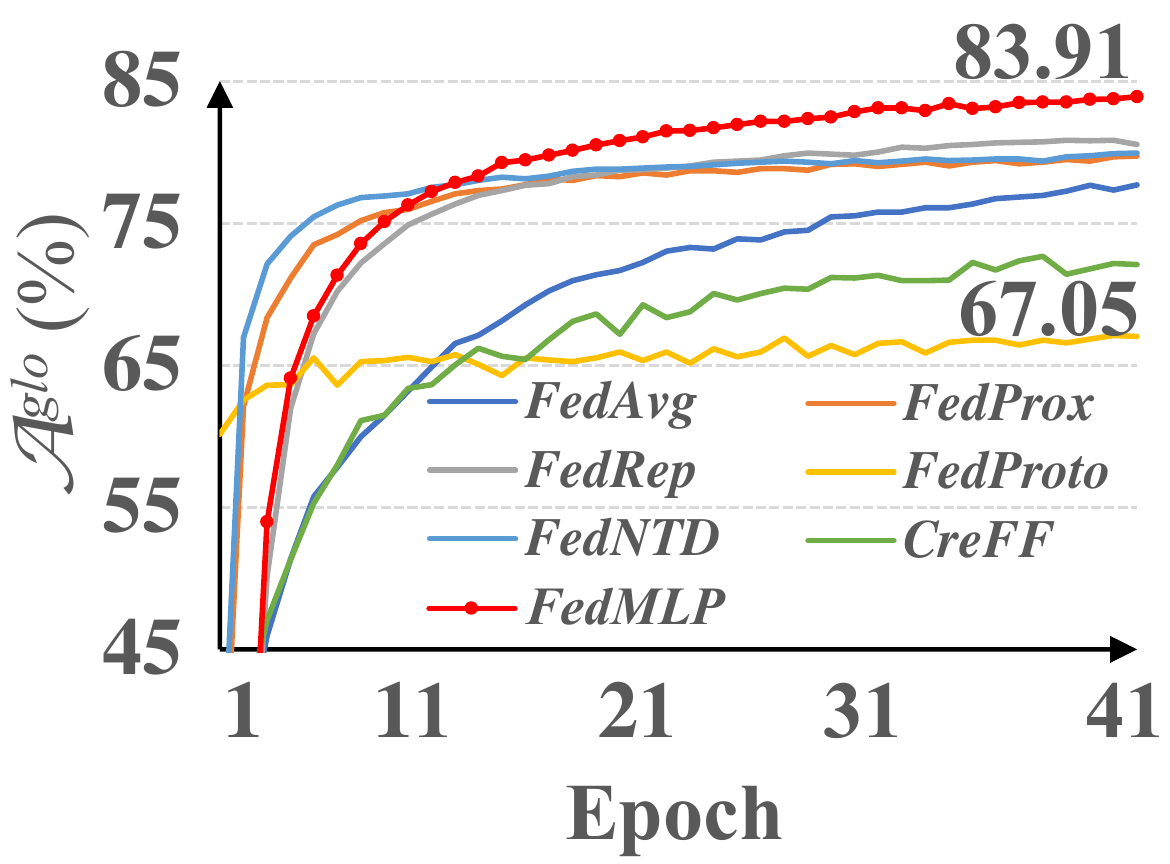}
        \caption{$\gamma$ = 1, $s$ = 4}
        \label{subfig:plot1}
    \end{subfigure}
    \begin{subfigure}{0.24\textwidth}
        \centering
        \includegraphics[width=\linewidth]{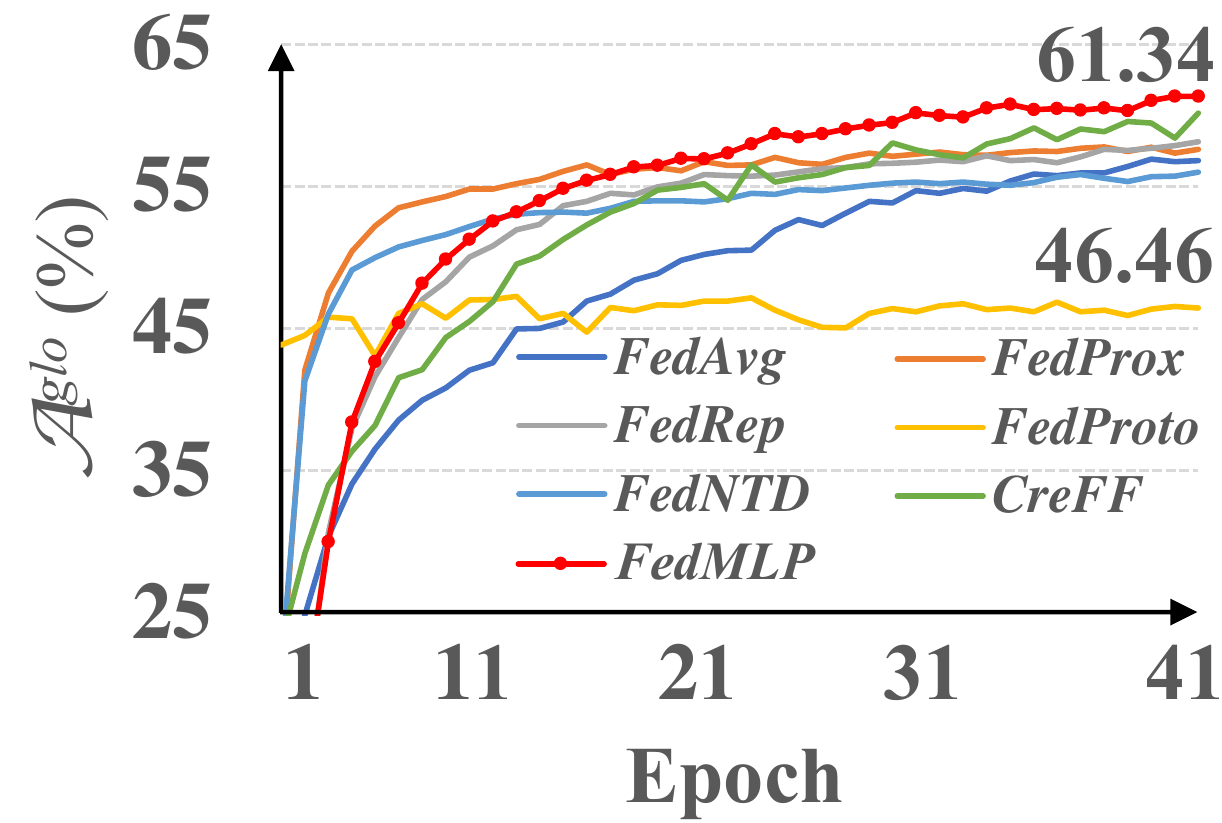}
        \caption{$\gamma$ = 0.05, $s$ = 4}
        \label{subfig:plot2}
    \end{subfigure}
    \caption{Comparison of $\mathcal{A}^{glo}$ at different global communication epochs for all methods on CIFAR-10 with $\gamma$ = 1, 0.05.}
    \label{fig:ifcom}
\end{figure}

\subsection{Comparison with State-of-the-Art}
We conduct a comparative analysis between the proposed FedMLP and other state-of-the-art methods on both local and global performance.
Results of the \textit{sharding} partition strategy of four benchmarks are summarized in Table~\ref{tab: com}. Our approach shows a notable superiority over the other FL methods. Specifically, in terms of $\mathcal{A}^{glo}$ on the CIFAR-100 dataset, our method performs 23.99\% higher than the recent FedNTD~\cite{lee2022preservation}. These results underscore the advantage of utilizing prototypes for training the generalized global model.
Furthermore, we found that the average results of all local models of the FedMLP are about the same as those of the global model. This observation provides compelling evidence for the capacity of the FedMLP to effectively train a universally applicable global model while maintaining the stability of local models.

The results of the \textit{DDA} partition strategy are presented in Table~\ref{tab:com2}.
The proposed FedMLP consistently achieves promising results, outperforming all baseline methods under various experimental settings. Notably, with greater heterogeneity when $\beta = 0.1$, FedMLP beats the state-of-the-art FedNTD~\cite{lee2022preservation} by 7.01\%.

\subsection{Model Analysis}
\noindent{\textbf{Global Concept Drift Analysis.} Figure~\ref{fig:ifcom} illustrates the $\mathcal{A}^{glo}$ for each communication epoch under two scenarios: $\gamma$ = 0.05 and $\gamma$ = 1.0. The results indicate that FedMLP exhibits faster convergence and more prominent results than other methods, despite the extreme imbalance in the global class distribution. Notably, the CreFF approach is designed to address federated global class imbalances but exhibits poor performance. Due to the incremental task within the clients, the local models tend to forget previously learned information, resulting in poor generalization of the trained balanced federated features.
The performance of the FedProto method remains static across epochs due to the inherent deviation between computed and expected prototypes for limited data.

\noindent{\textbf{Local Catastrophic Forgetting Analysis.}
In Figure~\ref{fig:task}, we compare the average accuracy of the local models after introducing a new task within the client. It can be seen that catastrophic forgetting in local models is inevitable due to the inconsistent distribution of local models across tasks. Compared to other methods, the proposed FedMLP exhibits a more gradual decay, indicating a smoother state of forgetting. FedNTD~\cite{lee2022preservation} shows promising results as it is specifically designed to counteract forgetting caused by diverse distributions.
Figures~\ref{fig:task}(b) and (d) illustrate
a similar decline in the accuracy of the minority class as new tasks are introduced. The proposed FedMLP also displays a more gradual attenuation trend compared to other methods, implying that classes with limited samples can achieve comparable performance to those with abundant samples.

\begin{figure}[tb]
    \centering
    \begin{subfigure}{0.23\textwidth}
        \centering
        \includegraphics[width=\linewidth]{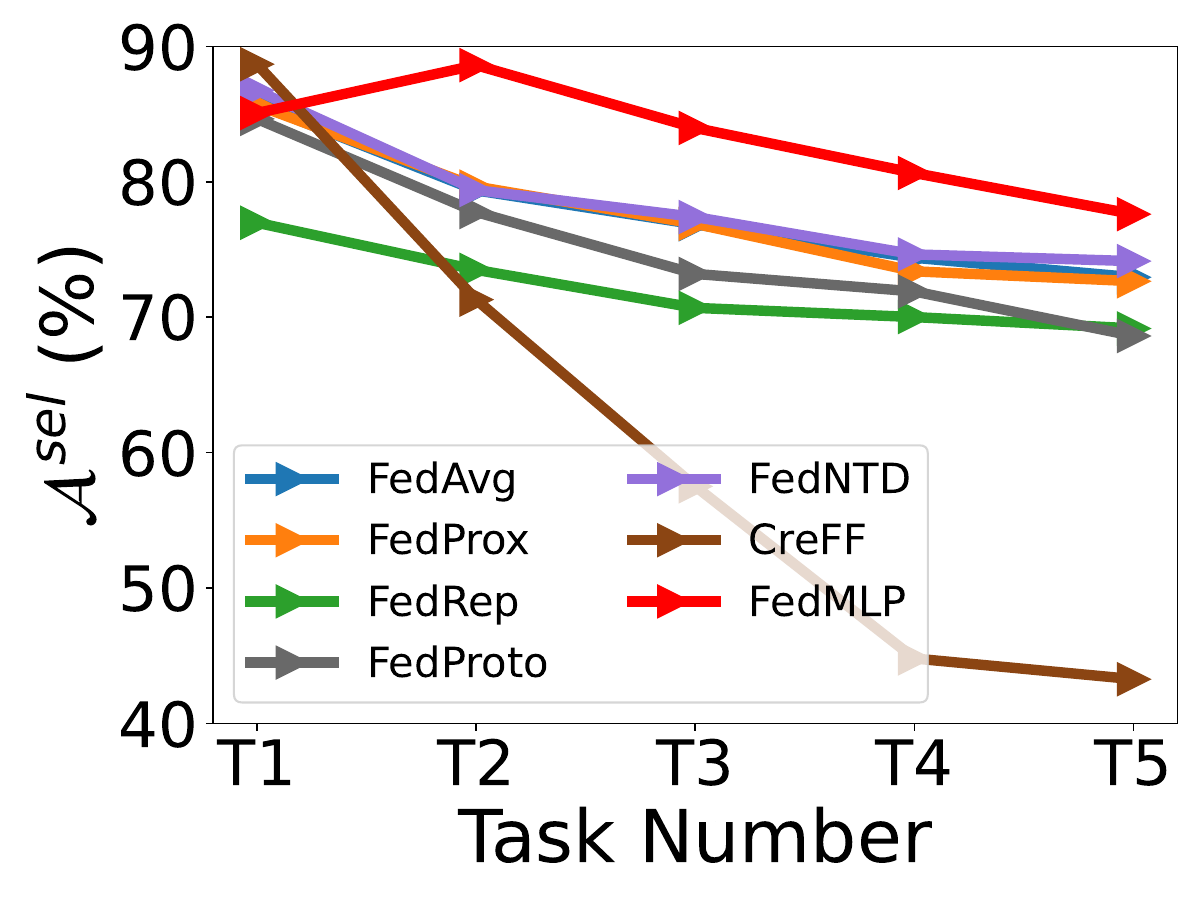}
        \caption{$\gamma$ = 0.5, $\beta$ = 0.1}
        \label{subfig:plot1}
    \end{subfigure}
    \begin{subfigure}{0.23\textwidth}
        \centering
        \includegraphics[width=\linewidth]{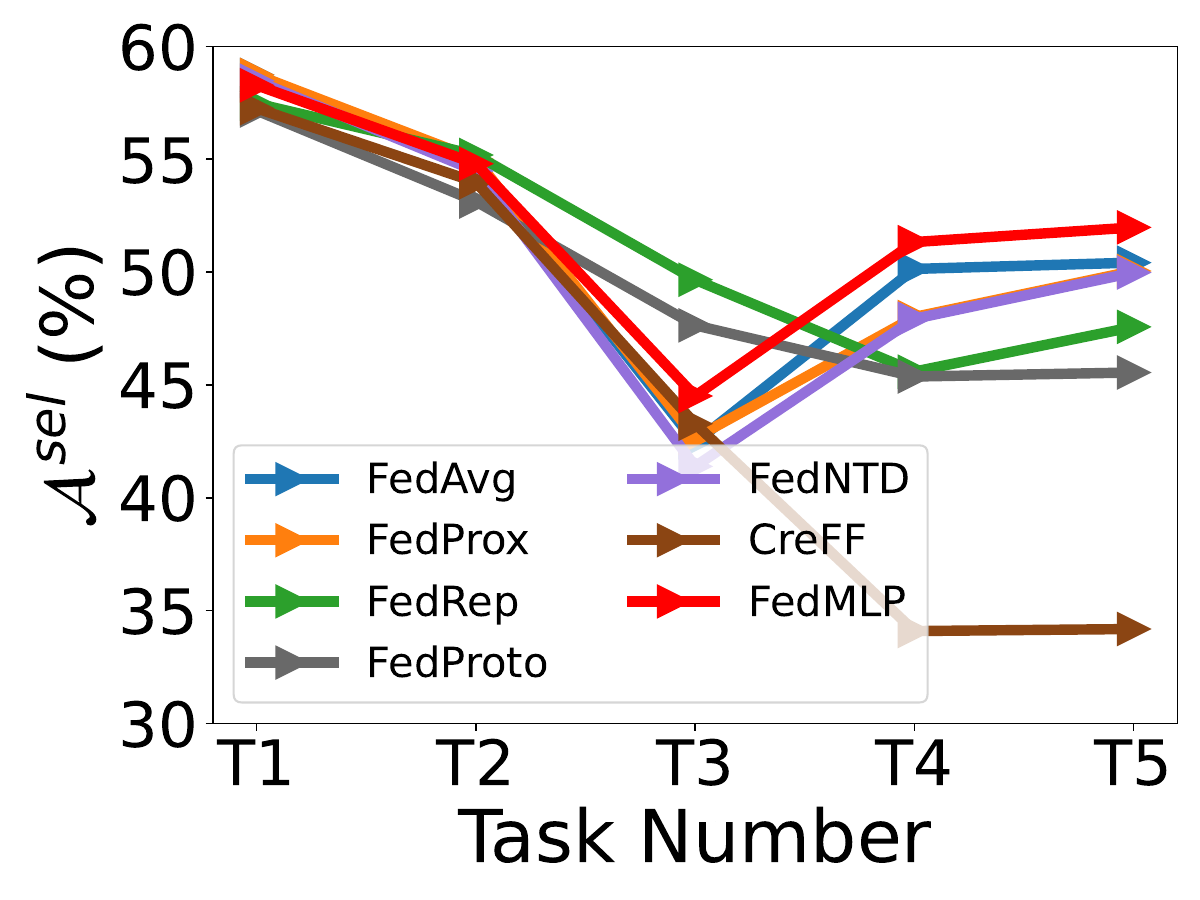}
        \caption{$\gamma$ = 0.5, $\beta$ = 0.1}
        \label{subfig:plot2}
    \end{subfigure}

    \begin{subfigure}{0.23\textwidth}
        \centering
        \includegraphics[width=\linewidth]{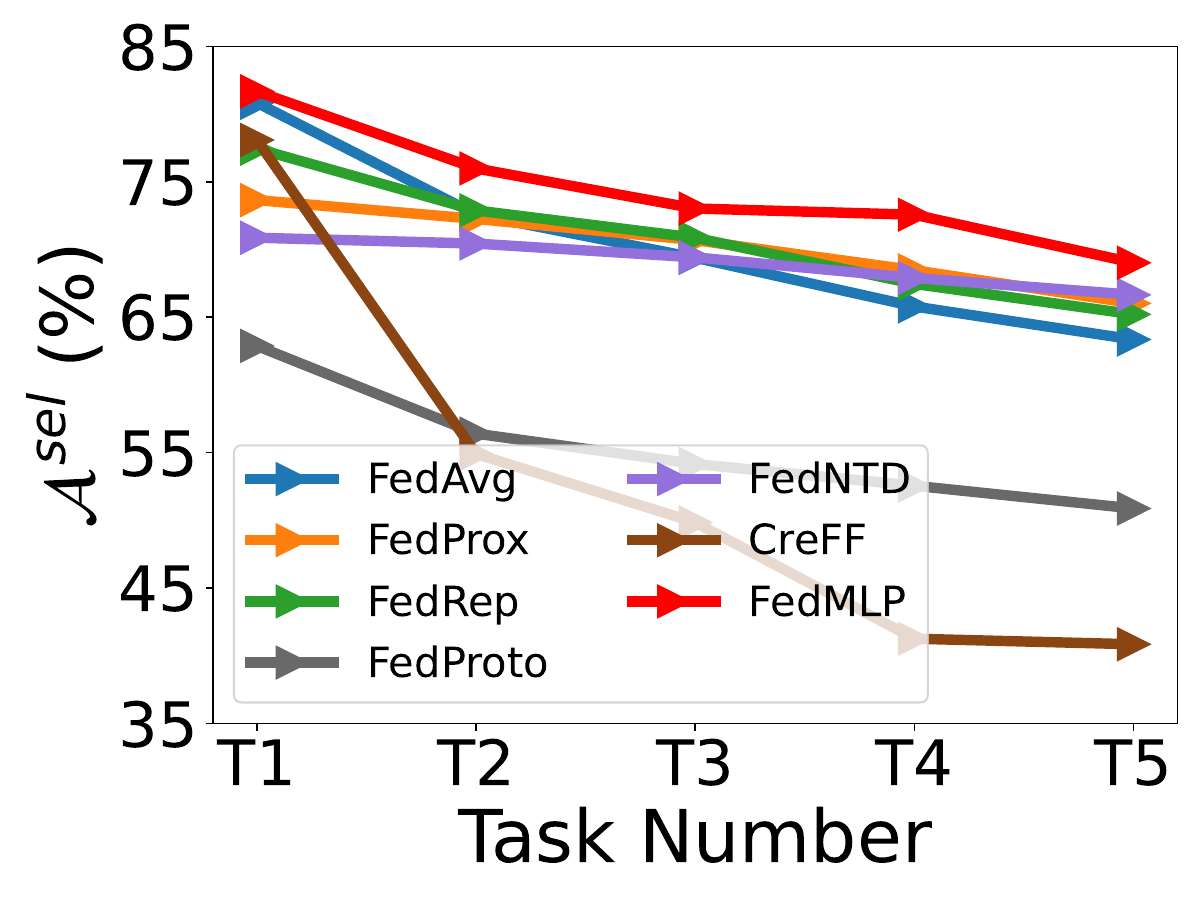}
        \caption{$\gamma$ = 0.5, $s$ = 4}
        \label{subfig:plot3}
    \end{subfigure}
    \begin{subfigure}{0.23\textwidth}
        \centering
        \includegraphics[width=\linewidth]{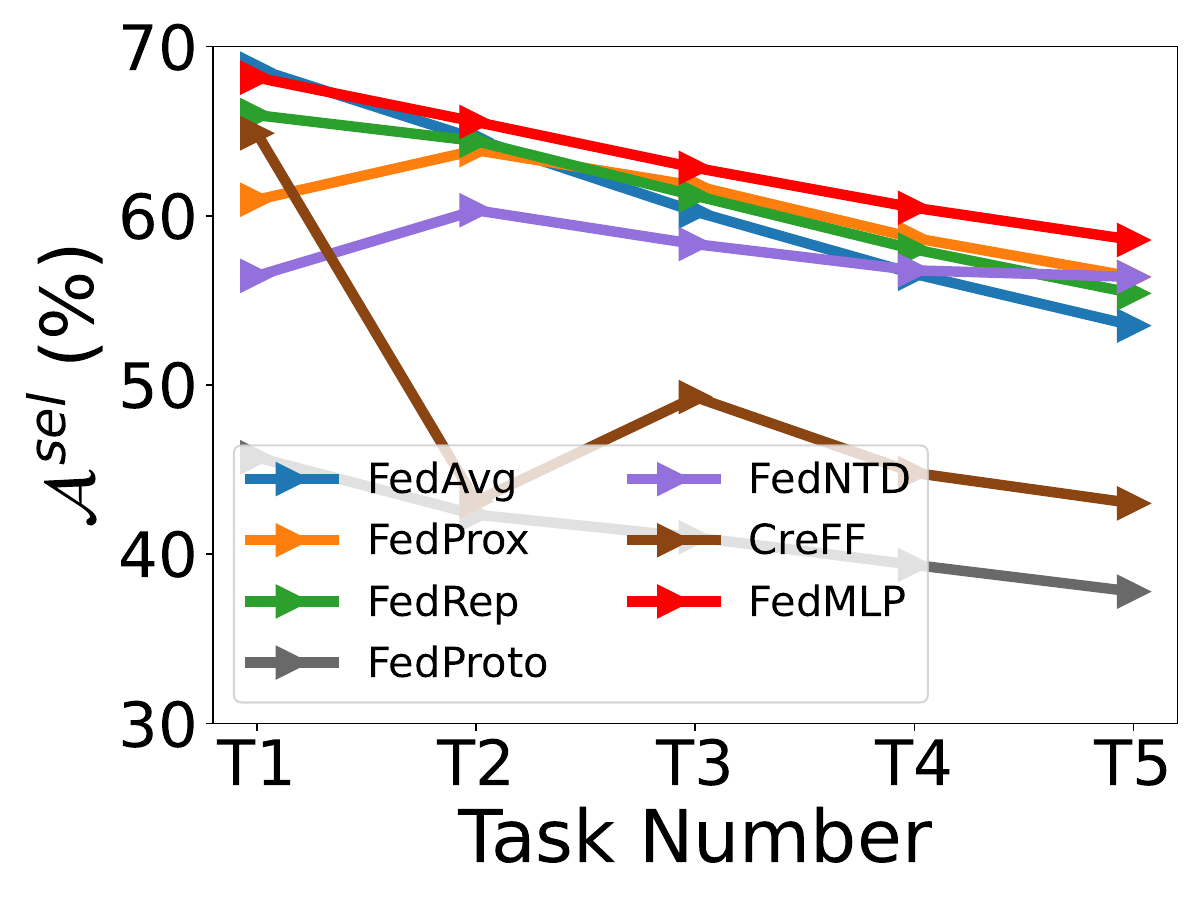}
        \caption{$\gamma$ = 0.5, $s$ = 4}
        \label{subfig:plot4}
    \end{subfigure}
    \caption{Comparison of $\mathcal{A}^{sel}$ on CIFAR-10 with the two non-iid partition strategies.}
    \label{fig:task}
\end{figure}

\noindent{\textbf{Ablation Studies.} The results of ablation studies on the CIFAR-10 with $\gamma$ = 0.5, $s$ = 4 are shown in Table~\ref{tab:ab}.
Symbols $\mathcal{A}^{loc-t}$ and $\mathcal{A}^{glo-t}$ denote the accuracy of the local and global models for minority classes, respectively.
Through the comparison of the $\mathcal{A}^{loc-t}$ and $\mathcal{A}^{glo-t}$ in different settings, it can be seen that the $\mathcal{L}_{S}$ used to provide supplementary generalization knowledge exerts a substantial influence on the training of auxiliary minority classes.
Compared with the proposed FedMLP, the performance of the baseline degrades significantly.
The experiments verify that all three objective losses are essential and complementary to each other for solving DHFL.

\begin{table}[t]
\label{ab}
\centering
\scalebox{0.75}{
\begin{tabular*}{11cm}{@{\extracolsep{\fill}}l|ccc|cccc}
\toprule
\textbf{Settings} &$\mathcal{L}_{P}$&$\mathcal{L}_{I}$&$\mathcal{L}_{S}$& $\mathcal{A}^{loc}$ & $\mathcal{A}^{loc-t}$ & $\mathcal{A}^{glo}$ & $\mathcal{A}^{glo-t}$ \\
\midrule
Baseline  &\XSolidBrush& \XSolidBrush & \XSolidBrush& 74.25 & 71.89& 74.72 & 71.99 \\
Baseline w/ $\mathcal{L}_{P}$  &\Checkmark & \XSolidBrush & \XSolidBrush &75.58 & 72.84  &  74.61 & 72.82 \\
Baseline w/ $\mathcal{L}_{I}$   &\XSolidBrush & \Checkmark & \XSolidBrush &75.63  & 73.27& 75.95 & 72.89 \\
Baseline w/ $\mathcal{L}_{S}$   &\XSolidBrush& \XSolidBrush & \Checkmark  &73.56 & 72.56 & 74.23 & 73.67 \\
\midrule
FedMLP w/o $\mathcal{L}_{P}$  &\XSolidBrush & \Checkmark & \Checkmark & 77.45  & 76.92  & 78.71  & 77.39 \\
FedMLP w/o $\mathcal{L}_{I}$ & \Checkmark & \XSolidBrush & \Checkmark& 76.64  & 77.12  & 76.05  & 78.05 \\
FedMLP w/o $\mathcal{L}_{S}$  &\Checkmark & \Checkmark & \XSolidBrush & 78.25  & 78.11  & 77.44  & 76.32\\
 FedMLP (Ours) &\Checkmark & \Checkmark &\Checkmark & 80.56 & 80.05 & 81.96 & 81.24 \\
\bottomrule
\end{tabular*}}
\caption{Ablation studies for the proposed FedMLP. }
\label{tab:ab}
\end{table}

\begin{figure}[tb]
\begin{center}
 \includegraphics[width=1\linewidth]{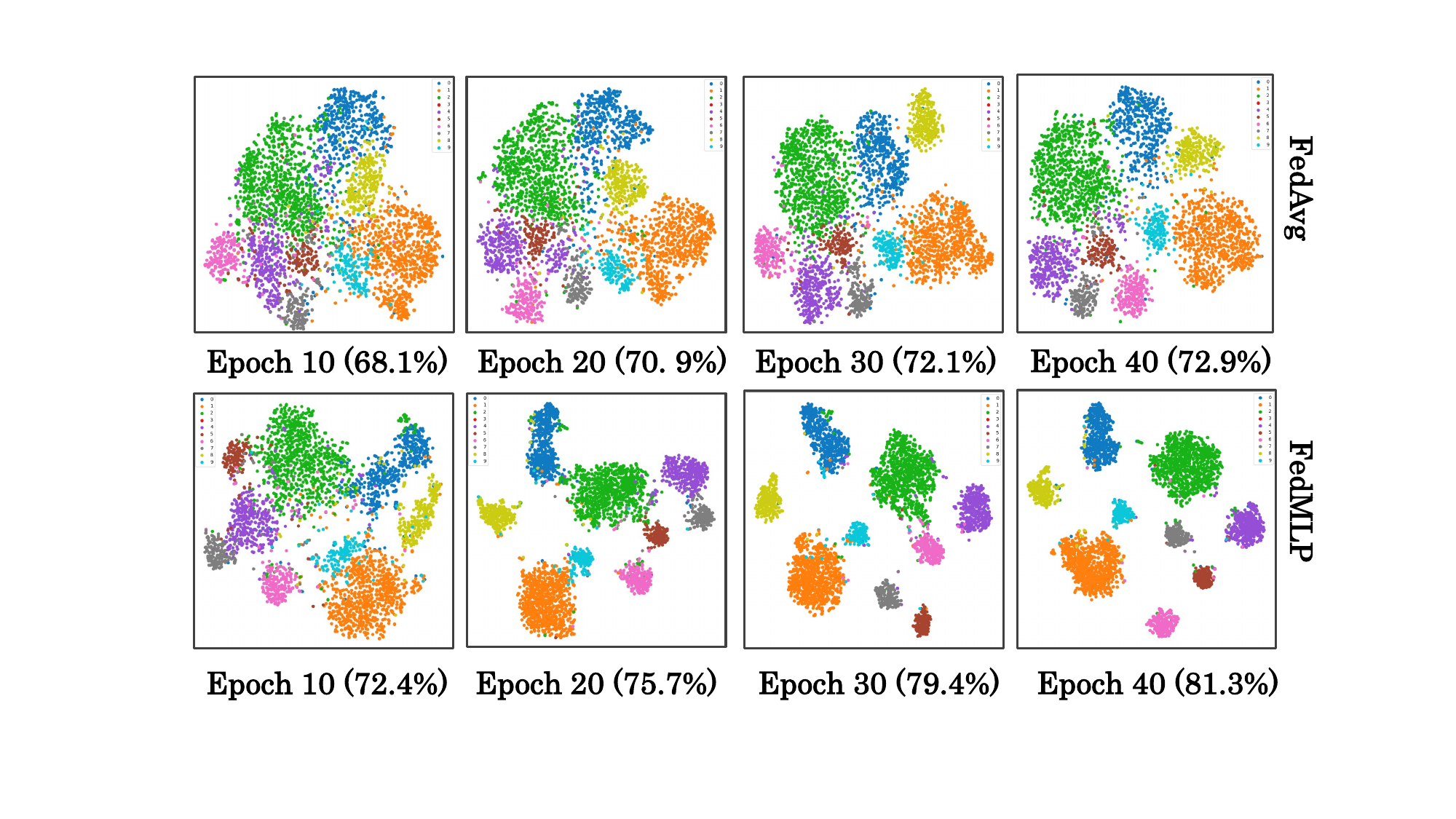}
\end{center}
   \caption{T-SNE visualizations of FedAvg~\cite{mcmahan2017communication} and FedMLP at different communication epochs on the CIFAR-10 dataset with $\gamma$ = 0.5, $s$ = 4.}
\label{fig:tsne}
\end{figure}
\noindent{\textbf{Feature Visualizations.} We present t-SNE visualization analyses between the proposed FedMLP and the notorious FedAvg at different communication epochs in Figure~\ref{fig:tsne}. Our proposed method demonstrates remarkable performance during the initial training phase, with distinct boundaries among classes indicating enhanced discriminatory capability. The results show that FedMLP is capable of learning a generalizable decision boundary even for minority classes with few samples, which alleviates the inherent global classification shift of long-tail class distribution.

\section{Conclusion}
In this paper, we introduce a novel task named Dynamic Heterogeneous Federated Learning (DHFL) and thoroughly investigate the associated issues, which encompass local catastrophic forgetting and global concept drift. To address these challenges, we present a novel Federated Multi-Level Prototypes (FedMLP) framework coupled with federated multi-level regularizations.
We conduct a comprehensive study on the DHFL using various experimental settings on four important benchmarks. Our proposed method significantly beats the recent state-of-the-art federated learning methods, while ablation studies demonstrate the effectiveness of each regularization.
FedMLP could effectively train a robust and generalized federated model with excellent results while preserving privacy and reducing communication overhead.

\bibliography{aaai24}

\begin{thebibliography}{38}
\providecommand{\natexlab}[1]{#1}

\bibitem[{Afonin and Karimireddy(2021)}]{afonin2021towards}
Afonin, A.; and Karimireddy, S.~P. 2021.
\newblock Towards model agnostic federated learning using knowledge
  distillation.
\newblock \emph{arXiv preprint arXiv:2110.15210}.

\bibitem[{Caldarola et~al.(2021)Caldarola, Mancini, Galasso, Ciccone,
  Rodol{\`a}, and Caputo}]{caldarola2021cluster}
Caldarola, D.; Mancini, M.; Galasso, F.; Ciccone, M.; Rodol{\`a}, E.; and
  Caputo, B. 2021.
\newblock Cluster-driven graph federated learning over multiple domains.
\newblock In \emph{Proceedings of the IEEE/CVF Conference on Computer Vision
  and Pattern Recognition}, 2749--2758.

\bibitem[{Caldas et~al.(2018)Caldas, Duddu, Wu, Li, Kone{\v{c}}n{\`y}, McMahan,
  Smith, and Talwalkar}]{caldas2018leaf}
Caldas, S.; Duddu, S. M.~K.; Wu, P.; Li, T.; Kone{\v{c}}n{\`y}, J.; McMahan,
  H.~B.; Smith, V.; and Talwalkar, A. 2018.
\newblock Leaf: A benchmark for federated settings.
\newblock \emph{arXiv preprint arXiv:1812.01097}.

\bibitem[{Cao et~al.(2019)Cao, Wei, Gaidon, Arechiga, and Ma}]{cao2019learning}
Cao, K.; Wei, C.; Gaidon, A.; Arechiga, N.; and Ma, T. 2019.
\newblock Learning imbalanced datasets with label-distribution-aware margin
  loss.
\newblock \emph{Advances in Neural Information Processing Systems}, 32.

\bibitem[{Cohen et~al.(2017)Cohen, Afshar, Tapson, and
  Van~Schaik}]{cohen2017emnist}
Cohen, G.; Afshar, S.; Tapson, J.; and Van~Schaik, A. 2017.
\newblock EMNIST: Extending MNIST to handwritten letters.
\newblock In \emph{2017 international joint conference on neural networks
  (IJCNN)}, 2921--2926. IEEE.

\bibitem[{Collins et~al.(2021)Collins, Hassani, Mokhtari, and
  Shakkottai}]{collins2021exploiting}
Collins, L.; Hassani, H.; Mokhtari, A.; and Shakkottai, S. 2021.
\newblock Exploiting shared representations for personalized federated
  learning.
\newblock In \emph{International Conference on Machine Learning}, 2089--2099.

\bibitem[{De~Boer et~al.(2005)De~Boer, Kroese, Mannor, and
  Rubinstein}]{de2005tutorial}
De~Boer, P.-T.; Kroese, D.~P.; Mannor, S.; and Rubinstein, R.~Y. 2005.
\newblock A tutorial on the cross-entropy method.
\newblock \emph{Annals of operations research}, 134: 19--67.

\bibitem[{Deng(2012)}]{deng2012mnist}
Deng, L. 2012.
\newblock The mnist database of handwritten digit images for machine learning
  research [best of the web].
\newblock \emph{IEEE signal processing magazine}, 29(6): 141--142.

\bibitem[{Dong et~al.(2022)Dong, Wang, Fang, Sun, Xu, Wang, and
  Zhu}]{dong2022federated}
Dong, J.; Wang, L.; Fang, Z.; Sun, G.; Xu, S.; Wang, X.; and Zhu, Q. 2022.
\newblock Federated class-incremental learning.
\newblock In \emph{Proceedings of the IEEE/CVF Conference on Computer Vision
  and Pattern Recognition}, 10164--10173.

\bibitem[{Durmus et~al.(2021)Durmus, Yue, Ramon, Matthew, Paul, and
  Venkatesh}]{durmus2021federated}
Durmus, A.~E.; Yue, Z.; Ramon, M.; Matthew, M.; Paul, W.; and Venkatesh, S.
  2021.
\newblock Federated Learning Based on Dynamic Regularization.
\newblock In \emph{International Conference on Learning Representations}.

\bibitem[{Gao et~al.(2022)Gao, Fu, Li, Chen, Xu, and Xu}]{gao2022feddc}
Gao, L.; Fu, H.; Li, L.; Chen, Y.; Xu, M.; and Xu, C.-Z. 2022.
\newblock Feddc: Federated learning with non-iid data via local drift
  decoupling and correction.
\newblock In \emph{Proceedings of the IEEE/CVF conference on computer vision
  and pattern recognition}, 10112--10121.

\bibitem[{Girshick(2015)}]{girshick2015fast}
Girshick, R. 2015.
\newblock Fast r-cnn.
\newblock In \emph{Proceedings of the IEEE international conference on computer
  vision}, 1440--1448.

\bibitem[{Huang et~al.(2023)Huang, Ye, Shi, Li, and Du}]{huang2023rethinking}
Huang, W.; Ye, M.; Shi, Z.; Li, H.; and Du, B. 2023.
\newblock Rethinking Federated Learning With Domain Shift: A Prototype View.
\newblock In \emph{Proceedings of the IEEE/CVF Conference on Computer Vision
  and Pattern Recognition}, 16312--16322.

\bibitem[{Kaiming~He and Sun(2016)}]{He2016Deep}
Kaiming~He, S.~R., Xiangyu~Zhang; and Sun, J. 2016.
\newblock Deep residual learning for image recognition.
\newblock \emph{Proceedings of the IEEE/CVF Conference on Computer Vision and
  Pattern Recognition}, 5: 770--778.

\bibitem[{Krizhevsky, Hinton et~al.(2009)}]{krizhevsky2009learning}
Krizhevsky, A.; Hinton, G.; et~al. 2009.
\newblock Learning multiple layers of features from tiny images.

\bibitem[{Krizhevsky, Sutskever, and Hinton(2012)}]{krizhevsky2012imagenet}
Krizhevsky, A.; Sutskever, I.; and Hinton, G.~E. 2012.
\newblock Imagenet classification with deep convolutional neural networks.
\newblock \emph{Advances in neural information processing systems}, 25.

\bibitem[{Lee et~al.(2022)Lee, Jeong, Shin, Bae, and Yun}]{lee2022preservation}
Lee, G.; Jeong, M.; Shin, Y.; Bae, S.; and Yun, S.-Y. 2022.
\newblock Preservation of the Global Knowledge by Not-True Distillation in
  Federated Learning.
\newblock \emph{Advances in Neural Information Processing Systems}.

\bibitem[{Li and Wang(2019)}]{li2019fedmd}
Li, D.; and Wang, J. 2019.
\newblock Fedmd: Heterogenous federated learning via model distillation.
\newblock \emph{arXiv preprint arXiv:1910.03581}.

\bibitem[{Li, He, and Song(2021)}]{li2021model}
Li, Q.; He, B.; and Song, D. 2021.
\newblock Model-contrastive federated learning.
\newblock In \emph{Proceedings of the IEEE/CVF conference on computer vision
  and pattern recognition}, 10713--10722.

\bibitem[{Li et~al.(2021)Li, Hu, Beirami, and Smith}]{li2021ditto}
Li, T.; Hu, S.; Beirami, A.; and Smith, V. 2021.
\newblock Ditto: Fair and robust federated learning through personalization.
\newblock In \emph{International Conference on Machine Learning}, 6357--6368.
  PMLR.

\bibitem[{Li et~al.(2020)Li, Sahu, Zaheer, Sanjabi, Talwalkar, and
  Smith}]{li2020federated}
Li, T.; Sahu, A.~K.; Zaheer, M.; Sanjabi, M.; Talwalkar, A.; and Smith, V.
  2020.
\newblock Federated optimization in heterogeneous networks.
\newblock \emph{Proceedings of Machine learning and systems}, 2: 429--450.

\bibitem[{Liang et~al.(2020)Liang, Liu, Ziyin, Allen, Auerbach, Brent,
  Salakhutdinov, and Morency}]{liang2020think}
Liang, P.~P.; Liu, T.; Ziyin, L.; Allen, N.~B.; Auerbach, R.~P.; Brent, D.;
  Salakhutdinov, R.; and Morency, L.-P. 2020.
\newblock Think locally, act globally: Federated learning with local and global
  representations.
\newblock \emph{arXiv preprint arXiv:2001.01523}.

\bibitem[{Long et~al.(2023)Long, Xie, Shen, Zhou, Wang, and
  Jiang}]{long2023multi}
Long, G.; Xie, M.; Shen, T.; Zhou, T.; Wang, X.; and Jiang, J. 2023.
\newblock Multi-center federated learning: clients clustering for better
  personalization.
\newblock \emph{World Wide Web}, 26(1): 481--500.

\bibitem[{Luo et~al.(2021)Luo, Chen, Hu, Zhang, Liang, and Feng}]{luo2021no}
Luo, M.; Chen, F.; Hu, D.; Zhang, Y.; Liang, J.; and Feng, J. 2021.
\newblock No fear of heterogeneity: Classifier calibration for federated
  learning with non-iid data.
\newblock \emph{Advances in Neural Information Processing Systems}, 34:
  5972--5984.

\bibitem[{Ma et~al.(2022)Ma, Xie, Wang, Chen, and Shou}]{ma2022continual}
Ma, Y.; Xie, Z.; Wang, J.; Chen, K.; and Shou, L. 2022.
\newblock Continual federated learning based on knowledge distillation.
\newblock In \emph{Proceedings of the Thirty-First International Joint
  Conference on Artificial Intelligence}, volume~3.

\bibitem[{McMahan et~al.(2017)McMahan, Moore, Ramage, Hampson, and
  y~Arcas}]{mcmahan2017communication}
McMahan, B.; Moore, E.; Ramage, D.; Hampson, S.; and y~Arcas, B.~A. 2017.
\newblock Communication-efficient learning of deep networks from decentralized
  data.
\newblock In \emph{Artificial Intelligence and Statistics}, 1273--1282.

\bibitem[{Mu et~al.(2023)Mu, Shen, Cheng, Geng, Fu, Zhang, and
  Zhang}]{mu2023fedproc}
Mu, X.; Shen, Y.; Cheng, K.; Geng, X.; Fu, J.; Zhang, T.; and Zhang, Z. 2023.
\newblock Fedproc: Prototypical contrastive federated learning on non-iid data.
\newblock \emph{Future Generation Computer Systems}, 143: 93--104.

\bibitem[{Sattler, M{\"u}ller, and Samek(2020)}]{sattler2020clustered}
Sattler, F.; M{\"u}ller, K.-R.; and Samek, W. 2020.
\newblock Clustered federated learning: Model-agnostic distributed multitask
  optimization under privacy constraints.
\newblock \emph{IEEE transactions on neural networks and learning systems},
  32(8): 3710--3722.

\bibitem[{Shang et~al.(2022{\natexlab{a}})Shang, Lu, Huang, and
  Wang}]{shang2022federated}
Shang, X.; Lu, Y.; Huang, G.; and Wang, H. 2022{\natexlab{a}}.
\newblock Federated learning on heterogeneous and long-tailed data via
  classifier re-training with federated features.
\newblock \emph{arXiv preprint arXiv:2204.13399}.

\bibitem[{Shang et~al.(2022{\natexlab{b}})Shang, Lu, Huang, and
  Wang}]{2022Federated}
Shang, X.; Lu, Y.; Huang, G.; and Wang, H. 2022{\natexlab{b}}.
\newblock Federated learning on heterogeneous and long-tailed data via
  classifier re-training with federated features.
\newblock \emph{International Joint Conference on Artificial Intelligence}.

\bibitem[{T~Dinh, Tran, and Nguyen(2020)}]{t2020personalized}
T~Dinh, C.; Tran, N.; and Nguyen, J. 2020.
\newblock Personalized federated learning with moreau envelopes.
\newblock \emph{Advances in Neural Information Processing Systems}, 33:
  21394--21405.

\bibitem[{Ta{\"\i}k, Mlika, and Cherkaoui(2022)}]{taik2022clustered}
Ta{\"\i}k, A.; Mlika, Z.; and Cherkaoui, S. 2022.
\newblock Clustered Vehicular Federated Learning: Process and Optimization.
\newblock \emph{IEEE Transactions on Intelligent Transportation Systems}.

\bibitem[{Tan et~al.(2022)Tan, Long, Liu, Zhou, Lu, Jiang, and
  Zhang}]{tan2022fedproto}
Tan, Y.; Long, G.; Liu, L.; Zhou, T.; Lu, Q.; Jiang, J.; and Zhang, C. 2022.
\newblock Fedproto: Federated prototype learning across heterogeneous clients.
\newblock In \emph{AAAI Conference on Artificial Intelligence}, volume~1,
  3--19.

\bibitem[{Yang et~al.(2022)Yang, Han, Liu, Liu, Wei, and
  Zhang}]{yang2022enhancing}
Yang, G.; Han, A.; Liu, X.; Liu, Y.; Wei, T.; and Zhang, Z. 2022.
\newblock Enhancing Semantic-Consistent Features and Transforming
  Discriminative Features for Generalized Zero-Shot Classifications.
\newblock \emph{Applied Sciences}, 12(24): 12642.

\bibitem[{Yang et~al.(2021)Yang, Zhang, Hao, Spell, and Carin}]{yang2021flop}
Yang, Q.; Zhang, J.; Hao, W.; Spell, G.~P.; and Carin, L. 2021.
\newblock Flop: Federated learning on medical datasets using partial networks.
\newblock In \emph{Proceedings of the 27th ACM SIGKDD Conference on Knowledge
  Discovery \& Data Mining}, 3845--3853.

\bibitem[{Yoon et~al.(2021)Yoon, Jeong, Lee, Yang, and
  Hwang}]{yoon2021federated}
Yoon, J.; Jeong, W.; Lee, G.; Yang, E.; and Hwang, S.~J. 2021.
\newblock Federated continual learning with weighted inter-client transfer.
\newblock In \emph{International Conference on Machine Learning}, 12073--12086.
  PMLR.

\bibitem[{Zhang et~al.(2022{\natexlab{a}})Zhang, Shen, Ding, Tao, and
  Duan}]{zhang2022fine}
Zhang, L.; Shen, L.; Ding, L.; Tao, D.; and Duan, L.-Y. 2022{\natexlab{a}}.
\newblock Fine-tuning global model via data-free knowledge distillation for
  non-iid federated learning.
\newblock In \emph{Proceedings of the IEEE/CVF Conference on Computer Vision
  and Pattern Recognition}, 10174--10183.

\bibitem[{Zhang et~al.(2022{\natexlab{b}})Zhang, Li, Ma, Gao, Li, and
  Lin}]{zhang2022residual}
Zhang, Z.; Li, X.; Ma, T.; Gao, Z.; Li, C.; and Lin, W. 2022{\natexlab{b}}.
\newblock Residual-Prototype Generating Network for Generalized Zero-Shot
  Learning.
\newblock \emph{Mathematics}, 10(19): 3587.

\end{thebibliography}

\end{document}